\pdfoutput=1
\documentclass[11pt]{article}
\usepackage[table,xcdraw]{xcolor}

\usepackage[]{acl}

\usepackage{times}
\usepackage{latexsym}
\usepackage[T1]{fontenc}
\usepackage[utf8]{inputenc}
\usepackage{microtype}
\usepackage{inconsolata}
\usepackage{graphicx}
\usepackage{multirow}
\usepackage{booktabs}
\usepackage{amsmath}
\usepackage[]{algorithm2e}
\usepackage{verbatim}
\usepackage{tcolorbox}
\tcbuselibrary{most}
\usepackage{colortbl}

\newcommand{\secvs}{\vspace{-1mm}}

\title{Style-Compress: An LLM-Based Prompt Compression Framework Considering Task-Specific Styles}

\author{Xiao Pu \\
  University of California, Santa Barbara \\ \texttt{xiao\_pu@ucsb.edu} \\ \\\AND
  Tianxing He$^*$ \\
  Tsinghua University \\
\texttt{hetianxing@mail.tsinghua.edu.cn} \\
  \\\And
  Xiaojun Wan$^*$ \\
  Peking University \\
  \texttt{wanxiaojun@pku.edu.cn} \\}

\begin{document}
\maketitle

\def\thefootnote{*}\footnotetext{These authors are co-corresponding authors.}\def\thefootnote{\arabic{footnote}}

\begin{abstract}
Prompt compression condenses contexts while maintaining their informativeness for different usage scenarios. It not only shortens the inference time and reduces computational costs during the usage of large language models, but also lowers expenses when using closed-source models. In a preliminary study, we discover that when instructing language models to compress prompts, different compression styles (e.g., extractive or abstractive) impact performance of compressed prompts on downstream tasks. Building on this insight, we propose Style-Compress, a lightweight framework that adapts a smaller language model to compress prompts for a larger model on a new task without additional training. Our approach iteratively generates and selects effective compressed prompts as task-specific demonstrations through style variation and in-context
learning, enabling smaller models to act as efficient compressors with task-specific examples.
Style-Compress outperforms two baseline compression models in four tasks: original prompt reconstruction, text summarization, multi-hop QA, and CoT reasoning. In addition, with only 10 samples and 100 queries for adaptation, prompts compressed by Style-Compress achieve performance on par with or better than original prompts at a compression ratio of 0.25 or 0.5.

\end{abstract}

\section{Introduction}
The rapid advancement of large language models has revolutionized natural language processing, enhancing both text understanding and generation capabilities \citep{10.1145/3649506,10.1145/3605943}. Despite these advancements, the increased size of these models has resulted in longer inference time, higher computational cost, and greater environmental impact. Additionally, the token-based fees associated with closed-source models such as GPT-4 \citep{openai2023gpt} and Claude \citep{claude} further emphasize the need for more efficient solutions.

These challenges highlight the necessity for research into efficient inference techniques, with prompt compression emerging as a promising approach to mitigate these issues by reducing the length of input prompts while still maintaining performance. Previous research has explored compressing contexts into soft prompts \citep{wingate-etal-2022-prompt, chevalier-etal-2023-adapting}, however this method is ineffective for black-box models. Other studies have focused on task-agnostic prompt compression by identifying and pruning redundant tokens using perplexity as an important metric \citep{li-etal-2023-compressing, jiang-etal-2023-llmlingua}. This task-agnostic paradigm naturally encounters difficulties in adjusting to special characteristics of end-tasks. On the other hand, compression methods tailored for certain tasks \citep{xu2023recomp,10535182} require additional training and are challenging to transfer to other tasks. Due to these challenges, we aim to propose a training-free prompt compression framework that can automatically adapt a generative prompt compressor to end tasks, with minimal data for adaption.

\begin{figure}[]
    \centering
    \includegraphics[width=\linewidth]{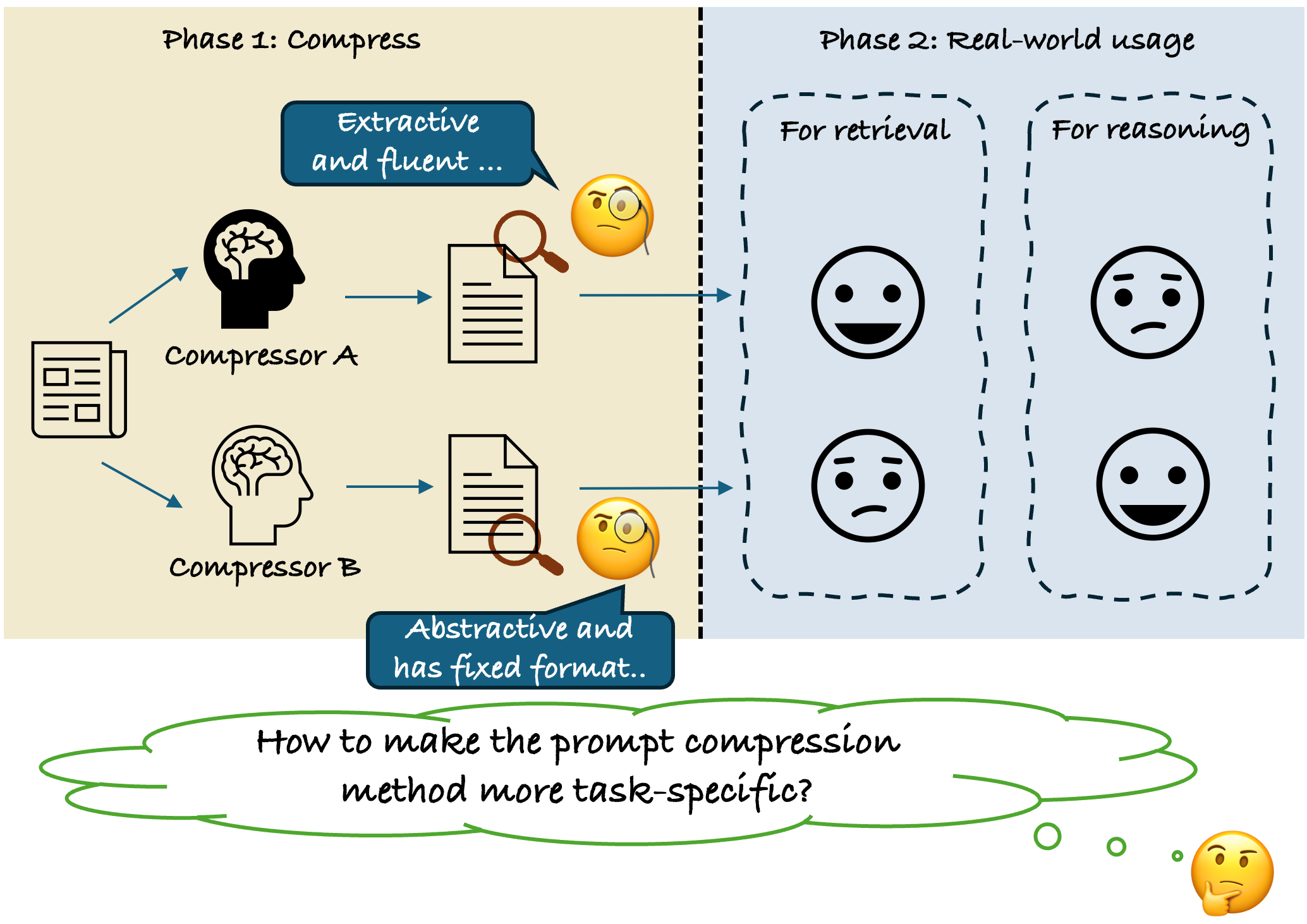}
    \caption{Our Motivation: Different usage scenarios benefit from different styles of prompt compression. For example, the retrieval QA task prefers compressed prompts that are extractive and fluent, whereas CoT reasoning favors compressed prompts that are abstractive and maintain a structured format. Therefore, we aim to automatically tailor prompt compression to specific downstream tasks from the perspective of styles.}
    \label{fig:motiv}
\end{figure}

Our key intuition is that different tasks prefer compressed prompts with different styles, e.g., extractive or abstractive (as shown in Figure \ref{fig:motiv}),  and the performance of prompt compression can be improved by iteratively learning to compress in the style beneficial to a specific task. Based on it, we propose a novel, lightweight framework, Style-Compress, which adapts smaller language models and compresses prompts for larger models in different downstream tasks without additional training. Our approach leverages the generative capabilities of smaller models and enhances their task-specific compression through iterative style variation and in-context learning. Specifically, our method involves iteratively generating and selecting representative compressions as demonstrations, enabling smaller language models to learn effective prompt compression tailored to specific tasks and evaluation models.

Our experimental results demonstrate that our method surpasses two baseline compression models across four tasks: original prompt reconstruction, text summarization, multi-hop QA, and CoT reasoning. Notably, prompts compressed by Style-Compress achieve performance that matches or exceeds that of original prompts at compression ratio of 0.25 or 0.5, with only 10 samples and 100 queries for adaptation. By improving prompt compression using smaller models, our research offers a valuable pathway towards more efficient and environmentally sustainable use of large language models.


\begin{table*}[h]
\centering
\resizebox{0.8\textwidth}{!}{
\begin{tabular}{@{}l|ll|ll|ll|l@{}}
\toprule
 &
  \multicolumn{2}{c|}{\textbf{Reconstruction}} &
  \multicolumn{2}{c|}{\textbf{Summarization}} &
  \multicolumn{2}{c|}{\textbf{Multi-hop QA}} &
  \textbf{Reasoning} \\ \cmidrule(l){2-8} 
\multirow{-2}{*}{\textbf{}} &
  \textbf{RougeL} &
  \textbf{BERTScore} &
  \textbf{RougeL} &
  \textbf{BERTScore} &
  \textbf{EM} &
  \textbf{F1} &
  \textbf{Acc} \\ \midrule
\textbf{vanilla} &
  \cellcolor[HTML]{FBD5D7}0.124 &
  \cellcolor[HTML]{FCF0F3}0.830 &
  \cellcolor[HTML]{FCFAFD}0.216 &
  \cellcolor[HTML]{FCFAFD}0.868 &
  \cellcolor[HTML]{FCE7EA}0.025 &
  \cellcolor[HTML]{F1F4FB}0.106 &
  \cellcolor[HTML]{FCDFE2}0.145 \\
\textbf{loc-begin} &
  \cellcolor[HTML]{F7F8FD}0.117 &
  \cellcolor[HTML]{F3F6FC}0.826 &
  \cellcolor[HTML]{FCF5F8}0.218 &
  \cellcolor[HTML]{FBFBFE}0.867 &
  \cellcolor[HTML]{FBBEC0}0.030 &
  \cellcolor[HTML]{F97E80}0.120 &
  \cellcolor[HTML]{FCF3F6}0.140 \\
\textbf{loc-mid} &
  \cellcolor[HTML]{FCF5F8}0.120 &
  \cellcolor[HTML]{FCEEF1}0.830 &
  \cellcolor[HTML]{F2F5FB}0.210 &
  \cellcolor[HTML]{FCFCFF}0.867 &
  \cellcolor[HTML]{FCE7EA}0.025 &
  \cellcolor[HTML]{FAB0B3}0.115 &
  \cellcolor[HTML]{FCF3F6}0.140 \\
\textbf{loc-end} &
  \cellcolor[HTML]{F8F9FD}0.118 &
  \cellcolor[HTML]{F2F5FB}0.826 &
  \cellcolor[HTML]{F1F4FB}0.210 &
  \cellcolor[HTML]{F8F9FD}0.866 &
  \cellcolor[HTML]{FCE7EA}0.025 &
  \cellcolor[HTML]{FBC5C7}0.113 &
  \cellcolor[HTML]{FBB8BA}0.155 \\
\textbf{loc-all} &
  \cellcolor[HTML]{FBCCCF}0.125 &
  \cellcolor[HTML]{FCFCFF}0.828 &
  \cellcolor[HTML]{E4EBF6}0.204 &
  \cellcolor[HTML]{F4F6FC}0.865 &
  \cellcolor[HTML]{EAEFF8}0.020 &
  \cellcolor[HTML]{FBCFD2}0.112 &
  \cellcolor[HTML]{9EBADE}0.110 \\
\textbf{abstractive} &
  \cellcolor[HTML]{FCE2E5}0.122 &
  \cellcolor[HTML]{FCE6E9}0.831 &
  \cellcolor[HTML]{FCDEE1}0.231 &
  \cellcolor[HTML]{FCDFE1}0.873 &
  \cellcolor[HTML]{A2BCDF}0.010 &
  \cellcolor[HTML]{FCF7FA}0.108 &
  \cellcolor[HTML]{F97D7F}0.170 \\
\textbf{extractive} &
  \cellcolor[HTML]{F2F5FB}0.116 &
  \cellcolor[HTML]{F5F7FC}0.827 &
  \cellcolor[HTML]{FCFCFF}0.214 &
  \cellcolor[HTML]{FCF6F9}0.868 &
  \cellcolor[HTML]{FBBEC0}0.030 &
  \cellcolor[HTML]{F8696B}0.122 &
  \cellcolor[HTML]{5A8AC6}0.090 \\
\textbf{readable} &
  \cellcolor[HTML]{FBD4D7}0.124 &
  \cellcolor[HTML]{FBFBFE}0.828 &
  \cellcolor[HTML]{FBFBFE}0.214 &
  \cellcolor[HTML]{F1F4FB}0.865 &
  \cellcolor[HTML]{FBBEC0}0.030 &
  \cellcolor[HTML]{FAA6A9}0.116 &
  \cellcolor[HTML]{FCDFE2}0.145 \\
\textbf{unreadable} &
  \cellcolor[HTML]{5A8AC6}0.068 &
  \cellcolor[HTML]{5D8CC7}0.792 &
  \cellcolor[HTML]{739CCF}0.157 &
  \cellcolor[HTML]{85A8D5}0.838 &
  \cellcolor[HTML]{FCE7EA}0.025 &
  \cellcolor[HTML]{F8F9FD}0.107 &
  \cellcolor[HTML]{F8696B}0.175 \\
\textbf{format-aware} &
  \cellcolor[HTML]{F5F7FC}0.117 &
  \cellcolor[HTML]{F9FAFE}0.828 &
  \cellcolor[HTML]{FAFAFE}0.213 &
  \cellcolor[HTML]{FBFBFE}0.867 &
  \cellcolor[HTML]{C6D6EC}0.015 &
  \cellcolor[HTML]{6793CA}0.086 &
  \cellcolor[HTML]{FBCCCE}0.150 \\
\textbf{for reconstruction} &
  \cellcolor[HTML]{F98284}0.134 &
  \cellcolor[HTML]{F98285}0.843 &
  \cellcolor[HTML]{F98789}0.279 &
  \cellcolor[HTML]{F97577}0.892 &
  \cellcolor[HTML]{5A8AC6}0.000 &
  \cellcolor[HTML]{8AABD6}0.091 &
  \cellcolor[HTML]{9EBADE}0.110 \\
\textbf{for summarization} &
  \cellcolor[HTML]{F8696B}0.137 &
  \cellcolor[HTML]{F97274}0.845 &
  \cellcolor[HTML]{F8696B}0.295 &
  \cellcolor[HTML]{F8696B}0.894 &
  \cellcolor[HTML]{A2BCDF}0.010 &
  \cellcolor[HTML]{5A8AC6}0.084 &
  \cellcolor[HTML]{7CA2D2}0.100 \\
\textbf{for qa} &
  \cellcolor[HTML]{FAABAD}0.129 &
  \cellcolor[HTML]{F8696B}0.846 &
  \cellcolor[HTML]{F97C7E}0.285 &
  \cellcolor[HTML]{F8696B}0.894 &
  \cellcolor[HTML]{F8696B}0.040 &
  \cellcolor[HTML]{FCE3E6}0.110 &
  \cellcolor[HTML]{F3F6FC}0.135 \\
\textbf{for reasoning} &
  \cellcolor[HTML]{F8F9FD}0.118 &
  \cellcolor[HTML]{FA9395}0.841 &
  \cellcolor[HTML]{F98C8E}0.276 &
  \cellcolor[HTML]{F97A7C}0.891 &
  \cellcolor[HTML]{EAEFF8}0.020 &
  \cellcolor[HTML]{F1F4FB}0.106 &
  \cellcolor[HTML]{E2EAF6}0.130 \\ \midrule
\textbf{baselines:} &
   &
   &
   &
   &
   &
   &
   \\
\textbf{selective-context} &
  \cellcolor[HTML]{EAEFF8}0.114 &
  \cellcolor[HTML]{5A8AC6}0.791 &
  \cellcolor[HTML]{5C8BC6}0.147 &
  \cellcolor[HTML]{5A8AC6}0.827 &
  \cellcolor[HTML]{C6D6EC}0.015 &
  \cellcolor[HTML]{9EBADE}0.094 &
  \cellcolor[HTML]{F3F6FC}0.135 \\
\textbf{LLM-Lingua} &
  \cellcolor[HTML]{B4C9E5}0.096 &
  \cellcolor[HTML]{6D97CC}0.796 &
  \cellcolor[HTML]{5A8AC6}0.146 &
  \cellcolor[HTML]{769ED0}0.834 &
  \cellcolor[HTML]{C6D6EC}0.015 &
  \cellcolor[HTML]{6793CA}0.086 &
  \cellcolor[HTML]{E2EAF6}0.130 \\ \bottomrule
\end{tabular}
}
\caption{Results of our preliminary study. Performance of prompting LLMs with different styles on four downstream tasks at a compression ratio of 10\%, compared with two compression methods (selective context and LLM-Lingua). A redder background indicates better performance, while a bluer background indicates worse performance. Different prompting styles yield varying performances across tasks, with no single style being optimal for all tasks.}
\label{tab:styles}
\end{table*}

\section{Preliminary Study: On the Role of Styles in Prompt Compression}

\label{sec:motiv}
In this section, we conduct a pilot experiment to explore the following research question: if we instruct LLMs to compress prompts, would the style of compression affect the performance of compressed prompts on various downstream tasks?

\textbf{Styles of Compression.} We consider five dimensions of styles \footnote{In this work, we define `style' as a broad and encompassing feature that distinguishes one piece of text from another.}, and instruct an LLM to do prompt compression with specific styles. We present the specific instructions to set these styles in Appendix \ref{appsec:instrution_style}:
\begin{itemize}
\secvs
    \item Location-aware: Focusing on different parts of the text to be compressed, such as the beginning, the middle, the end, or covering the entire text.
\secvs
    \item Abstractive or extractive: In light of the works in text summarization \citep{hahn2000challenges,carenini-cheung-2008-extractive}, we consider both abstractive and extractive styles of compression. Extractive style involves selecting words or sentence spans directly from the original text while abtractive style generates more creative and novel expressions.
    \secvs
    \item Readability: Does the compressed prompt need to be human-readable? It has been reported that GPT-4 can compress text in an unreadable way while still being able to reconstruct it \footnote{We first noticed it from this tweet: \url{https://x.com/VictorTaelin/status/1642664054912155648}}. We wonder if this trait also applies to smaller models.
    \secvs
    \item Format-aware: If there are specific formats in prompts, language models should retain such formats, which is called format-aware style.
    \secvs
    \item Task-aware: Can language models understand the type of compression required for different downstream tasks? To explore this, we explicitly inform the models of the task usage, e.g., "this is for summarization." 
\end{itemize}
\secvs

\textbf{Models and Tasks.} For starters, we use LLaMA-2 7B \citep{touvron2023llama} as both the compression and evaluation model, and larger models will be used for evaluation in later. We concatenate the instruction and the original prompt, and feed it to the compression model. We also incorporate two baseline compression models, Selective-Context \citep{li-etal-2023-compressing} and LLM-Lingua \citep{jiang-etal-2023-llmlingua}, for comparison. We incorporate four tasks: reconstruction, summarization, multi-hop QA and reasoning. Details of tasks, evaluation metrics and baseline models will be explained in Section 4.1.

\textbf{Results.}
The results of our preliminary study are shown in Table \ref{tab:styles}. We highlight two key findings: (1) Compared with baselines, even a vanilla prompt can achieve better prompt compression across all four tasks, highlighting the impressive compression performance of the LLM. (2) Style matters for prompt compression. Different styles lead to significantly varied performance on tasks. More importantly, no single style consistently outperforms the others across all tasks.

\textbf{Motivation.} According to our findings, certain compression styles in are preferred over others for specific tasks, as illustrated in Figure \ref{fig:motiv}. In light of this, we propose automatic adaptation of compression model to the task, from the perspective of styles. 
Our objective is to enable the compression LM to learn a specific style that enhances performance in the target task. We suggest that among a set of compressions with diverse styles derived from the same original prompt, those perform best on the task exhibit a task-relevant beneficial style. This specific style could be one defined by us (e.g., location-aware styles, abstractive or extractive), could be a blend or combination of some predefined styles, or even beyond them. We hypothesize that LMs can mimic these specific compression styles for target tasks when instructed with a few best-performing compressions as examples.


\begin{figure*}[h]
    \centering
    \includegraphics[width=\linewidth]{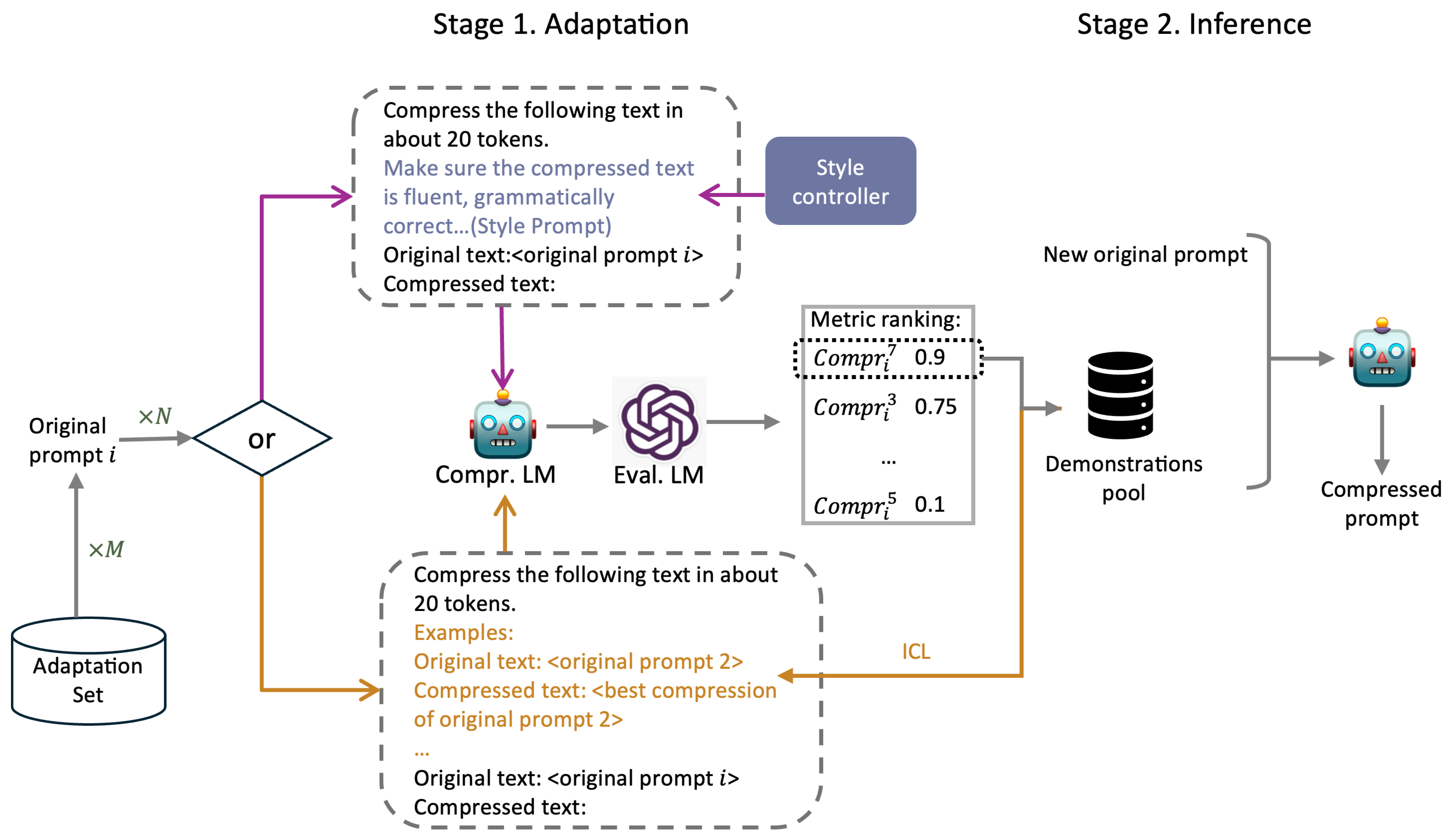}
    \caption{Illustration of Style-Compress. It adapts a smaller LM as a prompt compressor for a new task in two stages: In the first stage, the method iterates over $M$ prompts for adaptation, where for each prompt $N$ compressions are generated by (first) instructing the compression model using style variation (indicated by purple arrows) and (then) instructing the compressor with high-performing compressions as examples(indicated by orange arrows). The first stage iteratively generates a task-specific demonstration pool. We then use these examples to instruct the model in prompt compression during the inference stage.}
    \label{fig:method}
\end{figure*}

\section{Methodology}
\label{sec:method}

\subsection{Overview} Our goal is to adapt a smaller LM to compress prompts for a larger LM on a specific task. We refer to the smaller compression model as $LM_{compr}$ and the larger evaluation model as $LM_{eval}$. To achieve this, our method consists of two stages: (1) Adaptation: We generate prompts with diverse styles and then select the high-performing compressed prompts for this task. (2) Inference: We instruct $LM_{compr}$ with these selected ones as demonstrations to compress prompts for $LM_{eval}$. From these examples, the model can learn specific styles beneficial to the end task, improving task-specific prompt compression. Our proposed method is illustrated in Figure \ref{fig:method}.

\subsection{Adaptation} 

We generate and filter the high-performing compressions for a specific task within $M$ iterations. In each iteration $i$, a new original prompt $p_i$ is compressed by $LM_{compr}$ independently for $N$ times. We alternately use two methods to instruct $LM_{compr}$ in generating a variety of compressions, detailed in Section \ref{subsec:method3}. Then, $LM_{eval}$ generates outputs using the compressed prompt ${p'}_i^j$ (with $1 \leq j \leq n$) on the task. A metric value, ${m'}_i^j$, is computed to reflect the quality of ${p'}_i^j$ for this task. After generating and evaluating all $N$ compressions of the same original prompt, the best compression with respect to the metric, ${p'}_i^*$, and its \textit{comparative advantage} (CA) are added to the task-specific demonstrations pool. Depending on whether the best performance is compared to the worst or median, we have two variants of CA:

\small
\begin{align}
CA^\text{min}_i=max([{m'}_i^1,...,{m'}_i^n])-min([{m'}_i^1,...,{m'}_i^n]),\\
CA^\text{mid}_i = max([{m'}_i^1,...,{m'}_i^n])-mid([{m'}_i^1,...,{m'}_i^n]).
\end{align}
\normalsize

\subsection{Instruction methods for $LM_{compr}$}
\label{subsec:method3}
Two methods are designed to instruct $LM_{compr}$ in the first stage: style variation and in-context learning. The former diversifies styles, while the latter self-improves the quality of generation.

\textbf{Style Variation:} Style variation is designed to diversify the styles of compressed prompts. A style controller samples a specific style prompt from a set of human-written styles predefined by us. The selected style is added to the instruction for $LM_{compr}$ to compress the original prompt $p_i$, generating a compression, ${p'}_i^j$. When a warm-up ratio $R_\text{warm-up}$ is set for the style controller, it will perform random sampling during the initial 
$R_\text{warm-up} \cdot M$ iterations, then switch to weighted random sampling, based on the performance of different styles in past iterations, making styles with better performance more likely to be sampled.

\textbf{In-Context Learning:} For the ICL method, compressions with the highest CA in the demonstrations pool are selected as examples. By instructing $LM_{compr}$ with these task-specific high-performing examples, $LM_{compr}$ is able to mimic some specific styles beneficial to the target task from them.

For the same original prompt, we perform style variation for $N_{style}$ times and ICL for $N_{icl}$ times, thus $N = N_{style} + N_{icl}$.  Samples are generated and then ranked by CA. 

\subsection{Inference}
The first stage yields a task specific demonstrations pool including $M$ original prompts along with the best version of their compression and comparative advantages:[$({p}_1,{p'}_1^*,{CA}_1^*)$,...,$({p}_M,{p'}_M^*,{CA}_M^*)$]. We select $S$ pairs of original and compressed prompts with highest CAs from this pool as demonstrations to instruct $LM_{compr}$ to conduct task-specific prompt compression. The choice of S depends on the sample length of different task datasets. For each task dataset, we calculate the maximum number of demonstrations (including both the original and compressed prompts) that the compression model's context window can accommodate without affecting compression generation, to determine S.

\section{Experiment}
\label{sec:expr}
\subsection{Setting}
\textbf{Tasks and Datasets.}  We conduct experiments on four tasks: original prompt reconstruction using the BBC News dataset \citep{greene06icml}, text summarization with the CNN/DailyMail dataset \citep{see-etal-2017-get}, multi-hop QA using the HotpotQA dataset \citep{yang-etal-2018-hotpotqa}, and CoT reasoning with the GSM8k dataset \citep{cobbe2021gsm8k}. More specific details are provided in Appendix \ref{appsec:tasks}.

\textbf{Metrics.} We adopt Rouge-1, Rouge-2, Rouge-L \citep{lin-2004-rouge}, and BERTScore-F1 \citep{zhang2019bertscore} as metrics for the reconstruction and summarization tasks. We use the Evaluate, a library from huggingface to compute them.\footnote{https://github.com/huggingface/evaluate} For the reconstruction task, we use the original prompt as the reference, and for the summarization task, we use human-written summaries as the reference. We adopt EM and F1 as metrics for the multi-hop QA task and EM for the CoT reasoning task.

\textbf{Models.}  We use LLaMA-2-7B as the compression model. For the evaluation model, we choose the larger version of LLaMA-2, i.e. LLaMA-2 13B and a closed-source model, i.e. GPT-3.5. In the experiments, we compare Style-Compress with two baseline methods: Selective-Context \citep{li-etal-2023-compressing} and LLM-Lingua \citep{jiang-etal-2023-llmlingua}. It is also compared with a zero-shot approach where the language model is instructed to compress texts, which we refer to as the vanilla method.

\textbf{Details.} 
We experiment with three compression ratios: 0.1, 0.25 and 0.5. When instructing the language model to compress a prompt at a target compression ratio, we first compute the corresponding token count, which is the token count of original prompt multiplied by the set compression ratio, and include it in the instruction. For a fair comparison among all the methods, we truncate all the outputs to the target length. Results of our method are averaged using three random seeds. For hyper-parameters, we set $M=10, N=5, S=1$ for resconstruction and summarization, $S=2$ for QA and $S=3$ for reasoning. As for the setting of comparative advantage, we use ${CA}^{min}$ for reconstruction, summarization and multi-hop QA, and ${CA}^{mid}$ for the CoT reasoning task.
We conduct simple post-processing when zero-shot or few-shot prompting LLaMA-2 7B to compress text, because we find that the model tends to generate some redundant content, e.g., explanations and made-up examples.

\begin{table*}[h]
\centering
\resizebox{0.9\textwidth}{!}{%
\begin{tabular}{@{}lcccccccc@{}}
\toprule
\multicolumn{1}{l|}{\multirow{3}{*}{Methods}} &
  \multicolumn{8}{c}{Original Prompt Reconstruction} \\ \cmidrule(l){2-9} 
\multicolumn{1}{l|}{} &
  \multicolumn{4}{c|}{LLaMA-2 13B} &
  \multicolumn{4}{c}{GPT-3.5} \\
\multicolumn{1}{l|}{} &
  Rouge-1 &
  Rouge-2 &
  Rouge-L &
  \multicolumn{1}{c|}{BERTScore} &
  Rouge-1 &
  Rouge-2 &
  Rouge-L &
  BERTScore \\ \midrule
\multicolumn{9}{c}{\textbf{compression ratio = 0.1}} \\ \midrule
\multicolumn{1}{l|}{sc} &
  0.128 &
  0.020 &
  0.084 &
  \multicolumn{1}{c|}{0.694} &
  0.090 &
  0.021 &
  0.064 &
  0.800 \\
\multicolumn{1}{l|}{lingua} &
  0.103 &
  0.016 &
  0.070 &
  \multicolumn{1}{c|}{0.776} &
  0.136 &
  0.031 &
  0.094 &
  0.819 \\
\multicolumn{1}{l|}{vanilla} &
  0.202 &
  0.102 &
  0.138 &
  \multicolumn{1}{c|}{0.845} &
  0.181 &
  0.100 &
  0.133 &
  0.860 \\
\multicolumn{1}{l|}{ours} &
  \textbf{0.225} &
  \textbf{0.144} &
  \textbf{0.172} &
  \multicolumn{1}{c|}{\textbf{0.858}} &
  \textbf{0.205} &
  \textbf{0.145} &
  \textbf{0.165} &
  \textbf{0.872} \\ \midrule
\multicolumn{9}{c}{\textbf{compression ratio =  0.25}} \\ \midrule
\multicolumn{1}{l|}{sc} &
  0.253 &
  0.066 &
  0.151 &
  \multicolumn{1}{c|}{0.798} &
  0.196 &
  0.069 &
  0.130 &
  0.822 \\
\multicolumn{1}{l|}{lingua} &
  0.265 &
  0.075 &
  0.166 &
  \multicolumn{1}{c|}{0.814} &
  0.245 &
  0.076 &
  0.159 &
  0.840 \\
\multicolumn{1}{l|}{vanilla} &
  \textbf{0.326} &
  0.193 &
  0.234 &
  \multicolumn{1}{c|}{0.857} &
  0.319 &
  0.190 &
  0.239 &
  0.867 \\
\multicolumn{1}{l|}{ours} &
  0.324 &
  \textbf{0.213} &
  \textbf{0.240} &
  \multicolumn{1}{c|}{\textbf{0.865}} &
  \textbf{0.358} &
  \textbf{0.315} &
  \textbf{0.328} &
  \textbf{0.902} \\ \midrule
\multicolumn{9}{c}{\textbf{compression ratio = 0.5}} \\ \midrule
\multicolumn{1}{l|}{sc} &
  0.371 &
  0.146 &
  0.197 &
  \multicolumn{1}{c|}{0.849} &
  0.348 &
  0.173 &
  0.237 &
  0.856 \\
\multicolumn{1}{l|}{lingua} &
  0.377 &
  0.148 &
  0.223 &
  \multicolumn{1}{c|}{0.861} &
  0.424 &
  0.196 &
  0.303 &
  0.872 \\
\multicolumn{1}{l|}{vanilla} &
  0.372 &
  0.224 &
  0.268 &
  \multicolumn{1}{c|}{0.861} &
  0.370 &
  0.224 &
  0.277 &
  0.869 \\
\multicolumn{1}{l|}{ours} &
  \textbf{0.474} &
  \textbf{0.371} &
  \textbf{0.390} &
  \multicolumn{1}{c|}{\textbf{0.890}} &
  \textbf{0.516} &
  \textbf{0.453} &
  \textbf{0.475} &
  \textbf{0.919} \\ \bottomrule
\end{tabular}%
}
\caption{Results of compression methods on the reconstruction task. Here we abbreviate selective-context as sc and LLM-Lingua as lingua. }
\label{tab:res_reconstruct}
\end{table*}

\begin{table*}[htp]
\centering
\resizebox{0.9\textwidth}{!}{%
\begin{tabular}{@{}lllllllllllll@{}}
\toprule
\multicolumn{1}{l|}{\multirow{3}{*}{Methods}} &
  \multicolumn{8}{c|}{Summarization} &
  \multicolumn{4}{c}{Multi-hop QA} \\ \cmidrule(l){2-13} 
\multicolumn{1}{l|}{} &
  \multicolumn{4}{c|}{LLaMA-2 13B} &
  \multicolumn{4}{c|}{GPT-3.5} &
  \multicolumn{2}{c|}{LLaMA-2 13B} &
  \multicolumn{2}{c}{GPT-3.5} \\
\multicolumn{1}{l|}{} &
  \multicolumn{1}{c}{R-1} &
  \multicolumn{1}{c}{R-2} &
  \multicolumn{1}{c}{R-L} &
  \multicolumn{1}{c|}{BS} &
  \multicolumn{1}{c}{R-1} &
  \multicolumn{1}{c}{R-2} &
  \multicolumn{1}{c}{R-L} &
  \multicolumn{1}{c|}{BS} &
  \multicolumn{1}{c}{EM} &
  \multicolumn{1}{c|}{F1} &
  \multicolumn{1}{c}{EM} &
  \multicolumn{1}{c}{F1} \\ \midrule
\multicolumn{13}{c}{\textbf{compression ratio = 0.1}} \\ \midrule
\multicolumn{1}{l|}{sc} &
  0.108 &
  0.008 &
  0.084 &
  \multicolumn{1}{l|}{0.817} &
  0.128 &
  0.007 &
  0.096 &
  \multicolumn{1}{l|}{0.827} &
  0.005 &
  \multicolumn{1}{l|}{0.102} &
  \textbf{0.1} &
  \textbf{0.235} \\
\multicolumn{1}{l|}{lingua} &
  0.144 &
  0.017 &
  0.106 &
  \multicolumn{1}{l|}{0.831} &
  0.162 &
  0.014 &
  0.116 &
  \multicolumn{1}{l|}{0.837} &
  0.03 &
  \multicolumn{1}{l|}{0.107} &
  0.045 &
  0.187 \\
\multicolumn{1}{l|}{vanilla} &
  0.243 &
  0.078 &
  0.174 &
  \multicolumn{1}{l|}{0.860} &
  0.302 &
  0.103 &
  0.216 &
  \multicolumn{1}{l|}{0.871} &
  0.035 &
  \multicolumn{1}{l|}{0.116} &
  0.07 &
  0.202 \\
\multicolumn{1}{l|}{ours} &
  \textbf{0.270} &
  \textbf{0.089} &
  \textbf{0.191} &
  \multicolumn{1}{l|}{\textbf{0.866}} &
  \textbf{0.327} &
  \textbf{0.115} &
  \textbf{0.231} &
  \multicolumn{1}{l|}{\textbf{0.876}} &
  \textbf{0.05} &
  \multicolumn{1}{l|}{\textbf{0.135}} &
  0.098 &
  \textbf{0.235} \\ \midrule
\multicolumn{13}{c}{\textbf{compression ratio =  0.25}} \\ \midrule
\multicolumn{1}{l|}{sc} &
  0.133 &
  0.019 &
  0.098 &
  \multicolumn{1}{l|}{0.830} &
  0.161 &
  0.019 &
  0.118 &
  \multicolumn{1}{l|}{0.836} &
  0.02 &
  \multicolumn{1}{l|}{0.112} &
  0.08 &
  0.222 \\
\multicolumn{1}{l|}{lingua} &
  0.187 &
  0.034 &
  0.128 &
  \multicolumn{1}{l|}{0.846} &
  0.223 &
  0.039 &
  0.152 &
  \multicolumn{1}{l|}{0.853} &
  0.03 &
  \multicolumn{1}{l|}{0.134} &
  0.045 &
  0.186 \\
\multicolumn{1}{l|}{vanilla} &
  0.257 &
  0.086 &
  0.178 &
  \multicolumn{1}{l|}{0.864} &
  0.288 &
  0.096 &
  0.197 &
  \multicolumn{1}{l|}{0.869} &
  0.035 &
  \multicolumn{1}{l|}{0.125} &
  0.075 &
  0.214 \\
\multicolumn{1}{l|}{ours} &
  \textbf{0.288} &
  \textbf{0.101} &
  \textbf{0.200} &
  \multicolumn{1}{l|}{\textbf{0.870}} &
  \textbf{0.311} &
  \textbf{0.112} &
  \textbf{0.217} &
  \multicolumn{1}{l|}{\textbf{0.874}} &
  \textbf{0.055} &
  \multicolumn{1}{l|}{\textbf{0.146}} &
  \textbf{0.103} &
  \textbf{0.230} \\ \midrule
\multicolumn{13}{c}{\textbf{compression ratio = 0.5}} \\ \midrule
\multicolumn{1}{l|}{sc} &
  0.204 &
  0.054 &
  0.138 &
  \multicolumn{1}{l|}{0.852} &
  0.228 &
  0.060 &
  0.156 &
  \multicolumn{1}{l|}{0.858} &
  0.025 &
  \multicolumn{1}{l|}{0.129} &
  0.08 &
  0.223 \\
\multicolumn{1}{l|}{lingua} &
  0.221 &
  0.059 &
  0.150 &
  \multicolumn{1}{l|}{0.859} &
  0.254 &
  0.066 &
  0.171 &
  \multicolumn{1}{l|}{0.864} &
  0.02 &
  \multicolumn{1}{l|}{0.122} &
  0.045 &
  0.196 \\
\multicolumn{1}{l|}{vanilla} &
  0.262 &
  0.085 &
  0.178 &
  \multicolumn{1}{l|}{0.865} &
  0.284 &
  0.095 &
  0.193 &
  \multicolumn{1}{l|}{0.869} &
  0.035 &
  \multicolumn{1}{l|}{0.142} &
  0.05 &
  0.214 \\
\multicolumn{1}{l|}{ours} &
  \textbf{0.280} &
  \textbf{0.100} &
  \textbf{0.193} &
  \multicolumn{1}{l|}{\textbf{0.870}} &
  \textbf{0.297} &
  \textbf{0.107} &
  \textbf{0.205} &
  \multicolumn{1}{l|}{\textbf{0.872}} &
  \textbf{0.067} &
  \multicolumn{1}{l|}{\textbf{0.151}} &
  \textbf{0.09} &
  \textbf{0.239} \\ \midrule
\multicolumn{13}{c}{\textbf{no compression}} \\ \midrule
\multicolumn{1}{l|}{} &
  0.260 &
  0.101 &
  0.181 &
  \multicolumn{1}{l|}{0.867} &
  0.292 &
  0.106 &
  0.203 &
  \multicolumn{1}{l|}{0.872} &
  0.08 &
  \multicolumn{1}{l|}{0.207} &
  0.095 &
  0.265 \\ \bottomrule
\end{tabular}%
}
\caption{Results on the summarization and multi-hop QA tasks.}
\label{tab:res_summ_qa}
\end{table*}

\begin{table}[]
\centering
\resizebox{0.8\columnwidth}{!}{%
\begin{tabular}{@{}lllll@{}}
\toprule
\multicolumn{1}{c|}{}        & \multicolumn{2}{c|}{LLaMA-2 13B}                     & \multicolumn{2}{c}{GPT-3.5}    \\ \cmidrule(l){2-5} 
\multicolumn{1}{c|}{}        & \multicolumn{4}{c}{\cellcolor[HTML]{EFEFEF}0-shot}                                    \\ 
\multicolumn{1}{c|}{}        & \multicolumn{2}{c|}{0.195}                           & \multicolumn{2}{c}{0.55}       \\ \cmidrule(l){2-5} 
\multicolumn{1}{c|}{\multirow{-4}{*}{Methods}} &
  \cellcolor[HTML]{EFEFEF}1-shot &
  \multicolumn{1}{l|}{\cellcolor[HTML]{EFEFEF}3-shot} &
  \cellcolor[HTML]{EFEFEF}1-shot &
  \cellcolor[HTML]{EFEFEF}3-shot \\ \midrule
\multicolumn{5}{c}{\textbf{compression ratio = 0.1}}                                                                 \\ \midrule
\multicolumn{1}{l|}{sc}      & 0.09           & \multicolumn{1}{l|}{0.105}          & 0.625          & 0.395         \\
\multicolumn{1}{l|}{lingua}  & 0.115          & \multicolumn{1}{l|}{0.13}           & 0.62           & 0.575         \\
\multicolumn{1}{l|}{vanilla} & 0.135          & \multicolumn{1}{l|}{0.135}          & 0.615          & 0.61          \\
\multicolumn{1}{l|}{ours}    & \textbf{0.152} & \multicolumn{1}{l|}{\textbf{0.15}}  & \textbf{0.635}          & \textbf{0.647}          \\ \midrule
\multicolumn{5}{c}{\textbf{compression ratio = 0.25}}                                                                \\ \midrule
\multicolumn{1}{l|}{sc}      & 0.125          & \multicolumn{1}{l|}{0.115}          & 0.63           & 0.525         \\
\multicolumn{1}{l|}{lingua}  & 0.175          & \multicolumn{1}{l|}{0.14}           & 0.66           & 0.68          \\
\multicolumn{1}{l|}{vanilla} & 0.175          & \multicolumn{1}{l|}{0.18}           & 0.645          & 0.665         \\
\multicolumn{1}{l|}{ours}    & \textbf{0.19}  & \multicolumn{1}{l|}{\textbf{0.238}} & \textbf{0.665} & \textbf{0.73} \\ \midrule
\multicolumn{5}{c}{\textbf{compression ratio = 0.5}}                                                                 \\ \midrule
\multicolumn{1}{l|}{sc}      & 0.16          & \multicolumn{1}{l|}{0.195}          & 0.645          & 0.68          \\
\multicolumn{1}{l|}{lingua}  & 0.19           & \multicolumn{1}{l|}{0.19}           & 0.65           & 0.695         \\
\multicolumn{1}{l|}{vanilla} & 0.18           & \multicolumn{1}{l|}{0.195}          & 0.63           & 0.71          \\
\multicolumn{1}{l|}{ours}    & \textbf{0.193}  & \multicolumn{1}{l|}{\textbf{0.257}}   & \textbf{0.665} & \textbf{0.75} \\ \midrule
\multicolumn{5}{c}{\textbf{no compression}}                                                                          \\ \midrule
\multicolumn{1}{l|}{}        & 0.255          & \multicolumn{1}{l|}{0.335}          & 0.670          & 0.73         \\ \bottomrule
\end{tabular}%
}
\caption{Result on the CoT reasoning task.}
\label{tab:res_reason}
\end{table}

\subsection{Results}
Our main results on different tasks are presented in Table \ref{tab:res_reconstruct}, Table \ref{tab:res_summ_qa} and Table \ref{tab:res_reason}, respectively. Our method outperforms the baselines across all four tasks and enhances the prompt compression ability of LLMs. We include some examples of compressed prompts in Appendix \ref{appsec:examples}.

\textbf{Q1. How does our method compare to baselines?}
In the reconstruction, summarization and reasoning tasks, our method surpasses the baseline methods and vanilla approach at all three compression ratios, regardless of whether the evaluation model is GPT-3.5 or LLaMA-2 13B. For the multi-hop QA tasks, our method also demonstrates superior performance in most cases. The only exception is the QA task at a compression ratio of 0.1, where our method is slightly lower than selective-context in exact match (EM) but performs better in the F1 metric.
Our method's superior performance compared to the vanilla approach across all tasks demonstrate that our approach successfully enhances the prompt compression capabilities of smaller LMs.

\textbf{Q2. How well can LMs handle prompt compression?}
When directly instructing LLaMA-2 7B to compress text, its performance surpasses two baseline compression models in some cases, especially at a smaller compression ratio and in easier tasks like reconstruction and summarization. For instance, in the reconstruction task, the vanilla method (i.e., zero-shot instructing the model to compress) outperforms both selective-context and LLM-Lingua at ratios of 0.1 and 0.25. However, at a compression ratio of 0.5, the baselines start to overtake. For more challenging tasks like multi-hop and reasoning, the performance of the vanilla method and the baselines are comparable, occasionally being slightly higher or lower. Our results show that a smaller LM as LLaMA-2 7B has some ability to compress prompts, which aligns with our previous finding on a smaller evaluation model in Section \ref{sec:motiv}. Besides, it is quite feasible to use a smaller LM to compress prompts for a larger LM, e.g., GPT-3.5 and LLaMA-2 13B. 


\textbf{Q3. How much information is lost after compression?}
Except for the reconstruction task, where the original prompts serve as references, we compare performance of  compressed prompts with the original ones to analyse how much information is lost after compression. From the results we find that the performance drop is generally acceptable if not compressed excessively. In the summarization task, prompts compressed by our method perform better than original prompts in almost all metrics across all settings, even at an extreme compression ratio of 0.1. In the multi-hop QA task, when compressed using our method at a compression ratio of 0.5, the metrics is only slightly lower than using prompts without compression. Results from the CoT reasoning task align with this finding. When we perform 3-shot CoT prompting, with GPT-3.5 as the evaluation model, the accuracy without compression is 67\%, but after our compression at a ratio of 0.25, the accuracy still reaches 66.5\%. In some cases, performance even improves after compression. For example, in the 3-shot CoT reasoning task with a compression ratio of 0.5 and GPT-3.5 as the evaluation model, the accuracy after compression is 2\% higher than no compression. These results highlight the great potential of cost saving from prompt compression.

\subsection{Ablation Study and Parameter Analysis}
Here we explore how different settings in our proposed method would affect the performance of prompt compression. 
The default settings of our method in this section are as follows: The compression and evaluation models are all LLaMA-2 7B. Iteration count $M$ is set to 10, $N_{style}$ and $N_{icl}$ are both set to 5. We exclude warming-up in the style variation module. Due to space constraints, we only present the most important part of our ablation study, focusing on the effects of style variation and ICL for generating and selecting examples here. The analysis of the effects of comparative advantage, iteration count and warming-up in style controller is included in Appendix \ref{appsec:ablation}.

\textbf{Effects of style variation and ICL for generating and selecting examples. } To examine whether the two designed modules, style variation and In-Context Learning, are necessary for the adaptation stage in our method, we remove them separately and test the performance. As shown in Table \ref{tab:ablation_modules}, we observe a decline in accuracy in both cases. On the CoT reasoning task, accuracy of answers drops by 1.1\% without ICL and 2.8\% without styles variation. The reconstruction task also shows poorer performance when these modules are excluded.

\begin{table}[]
\resizebox{\columnwidth}{!}{%
\begin{tabular}{@{}l|l|llll@{}}
\toprule
\multicolumn{1}{c|}{\multirow{2}{*}{}} &
  \multicolumn{1}{c|}{Reasoning} &
  \multicolumn{4}{c}{Reconstruction} \\ \cmidrule(l){2-6} 
\multicolumn{1}{c|}{} &
  \multicolumn{1}{c|}{Acc.} &
  \multicolumn{1}{c}{Rouge1} &
  \multicolumn{1}{c}{Rouge2} &
  \multicolumn{1}{c}{RougeL} &
  \multicolumn{1}{c}{BS} \\ \midrule
Ours         & \textbf{0.12}  & \textbf{0.372} & \textbf{0.274} &\textbf{0.297} & \textbf{0.88}  \\
- w/o Styles & 0.092 & 0.344 & 0.244 & 0.271 & 0.875 \\
- w/o ICL    & 0.109 & 0.33  & 0.228 & 0.251 & 0.875 \\ \bottomrule
\end{tabular}%
}
\caption{Ablation study of style variation and ICL.}
\label{tab:ablation_modules}
\end{table}

\begin{table*}[h]
\resizebox{\linewidth}{!}{%
\begin{tabular}{@{}l|llll|llll|ll|c@{}}
\toprule
                                & \multicolumn{4}{c|}{Reconstruction} & \multicolumn{4}{c|}{Summarization}     & \multicolumn{2}{c|}{Multi-hop QA} & CoT Reasoning        \\
 &
  \multicolumn{1}{c}{R-1} &
  \multicolumn{1}{c}{R-2} &
  \multicolumn{1}{c}{R-L} &
  \multicolumn{1}{c|}{BS} &
  \multicolumn{1}{c}{R-1} &
  \multicolumn{1}{c}{R-2} &
  \multicolumn{1}{c}{R-L} &
  \multicolumn{1}{c|}{BS} &
  \multicolumn{1}{c}{EM} &
  \multicolumn{1}{c|}{F1} &
  Acc. \\ \midrule
Ours &
  \textbf{0.372} &
  \textbf{0.274} &
  \textbf{0.297} &
  \textbf{0.88} &
  \textbf{0.289} &
  \textbf{0.104} &
  \textbf{0.201} &
  \textbf{0.871} &
  \textbf{0.045} &
  \textbf{0.138} &
  \textbf{0.12} \\
Ours w/o ICL                    & 0.33    & 0.228   & 0.251  & 0.875  & \textbf{0.289} & 0.103 & 0.196 & 0.87  & \textbf{0.045}       & 0.126      & 0.109                \\
\multirow{2}{*}{BestStyle-CA}   & 0.334   & 0.196   & 0.237  & 0.862  & 0.274          & 0.091 & 0.186 & 0.867 & 0.03                 & 0.129      & 0.095                \\
                                & \multicolumn{4}{c|}{(format)}       & \multicolumn{4}{c|}{(loc-end)}         & \multicolumn{2}{c|}{(extractive)} & (loc-all,unreadable) \\
\multirow{2}{*}{BestStyle-Stat} & 0.331   & 0.182   & 0.229  & 0.859  & 0.278          & 0.102 & 0.191 & 0.869 & 0.005                & 0.088      & 0.105                \\
                                & \multicolumn{4}{c|}{(loc-all)}      & \multicolumn{4}{c|}{(extractive)}      & \multicolumn{2}{c|}{(unreadable)} & (format)             \\ \bottomrule
\end{tabular}%
}
\caption{Results of the discussion part. For the BestStyle methods, we place the selected best style in the bracket below the corresponding numbers. Compared with two BestStyle methods, our method has better performance across all four tasks. Even without ICL when generating and selecting demonstrations, our method still outperforms the BestStyles.}
\label{tab:discussion}
\end{table*}
\subsection{Discussion}
\textbf{What styles has the compressor learned?}
Our method involves two stages: first, we generate and select high-performing compressions as task-specific demonstrations. Then, we use these demonstrations to guide the LM in compressing prompts. As we propose in Section \ref{sec:motiv}, we believe that by leveraging high-performing compressions, the compressor can learn a task-specific style that is particularly beneficial.

To investigate what constitutes this "specific style" and how it differs from predefined styles, we compare our approach with an alternative: instead of using demonstrations, multiple predefined styles are established, and the best one for the task is identified and directly applied to guide the compressor during inference.

We specifically compare our method against two approaches that rely on selecting the best predefined style:

\secvs
\begin{itemize}
    \item BestStyle-CA: We maintain the comparative advantage setting of our method. We compare different compression styles for the same original prompts and record the best compression based on its comparative advantage. During evaluation, we select the style with the highest comparative advantage as the best style.
    \secvs
    \item BestStyle-Stat: For each original prompt, we conduct pairwise comparisons of the compression styles based on specific metrics. Finally, we choose the style with the highest win rate as the best style.
\end{itemize}

The results are shown in Table \ref{tab:discussion}. Our method outperforms both BestStyle methods across all four tasks, demonstrating its effectiveness. Furthermore, we find that even without ICL in the first stage, our method still generally surpasses the two BestStyle methods. It means that the specific style compressor learned is not identical to styles predefined by us. Instead, the learned specific style may incorporate different predefined styles or introduce new, task-beneficial elements due to the inherent randomness in the inference process of language models, thereby becoming a better and more precise compression style tailored to the target task.

\textbf{How generalizable are the compressions across LMs?} To explore this, we select the reconstruction task, set the compression model to LLaMA2-7B or GPT-3.5, and test the performance of compressions obtained from different models applied to other models.
We find a small-to-large generalization phenomenon of compressions within the same model family, as shown in Table \ref{tab:general1}. Especially, when applying compressions from LLaMA-2 7B to LLaMA-2 13B, the performance is better than LLaMA-2 13B's own compressions. However, this strong generalization ability is not observed in the reverse direction. We notice a similar phenomenon between GPT3.5 and GPT 4. Across different model families, we find that GPT-3.5 generalizes relatively well to LLaMA-2 13B, no matter the compression model is LLaMA-2 7B or GPT-3.5, as shown in Table \ref{tab:general2}.

\begin{table}[t]
\resizebox{\linewidth}{!}{%
\begin{tabular}{@{}ll|ll@{}}
\toprule
\textbf{From} & \textbf{To} & \textbf{Rouge-L} & \textbf{BERTScore} \\ \midrule
LLaMA-2 7B    & LLaMA-2 13B & \textbf{0.306}   & \textbf{0.881}     \\
LLaMA-2 13B   & LLaMA-2 13B & 0.240            & 0.865              \\
LLaMA-2 13B   & LLaMA-2 7B  & 0.252            & 0.866              \\
LLaMA-2 7B    & LLaMA-2 7B  & 0.297            & 0.880              \\ \midrule
GPT-3.5       & GPT-4       & \textbf{0.279}   & \textbf{0.889}     \\
GPT-4         & GPT-4       & 0.262            & 0.888              \\
GPT-4         & GPT-3.5     & \textbf{0.337}   & 0.899              \\
GPT-3.5       & GPT-3.5     & 0.328            & 0.902              \\ \bottomrule
\end{tabular}
}
\caption{Generalizability of compressions within the same model family in the reconstruction task when the compression model is set to LLaMA-2 7B. "From" indicates the LLM used as the evaluation model to obtain the compressions, and "To" indicates the model these compressions are tested on. Metrics are marked in bold if the performance of compressions from another model is better than the model's own compressions.}
\label{tab:general1}
\end{table}

\begin{table}[t]
\resizebox{\linewidth}{!}{%
\begin{tabular}{@{}l|ll|ll@{}}
\toprule
\textbf{Compressor}     & \textbf{From} & \textbf{To} & \textbf{R-L} & \textbf{B-S} \\ \midrule
\multirow{4}{*}{LLaMA-2 7B} & LLaMA-2 13B   & GPT-3.5     & 0.262            & 0.874              \\
                            & GPT-3.5       & GPT-3.5     & 0.328            & 0.902              \\
                            & GPT-3.5       & LLaMA-2 13B & \textbf{0.297}   & \textbf{0.872}     \\
                            & LLaMA-2 13B   & LLaMA-2 13B & 0.240            & 0.865              \\ \midrule
\multirow{4}{*}{GPT-3.5}    & LLaMA-2 13B   & GPT-3.5     & 0.195            & 0.861              \\
                            & GPT-3.5       & GPT-3.5     & 0.205            & 0.863              \\
                            & GPT-3.5       & LLaMA-2 13B & \textbf{0.297}   & \textbf{0.872}     \\
                            & LLaMA-2 13B   & LLaMA-2 13B & 0.182            & 0.855              \\ \bottomrule
\end{tabular}
}
\caption{Generalizability of compressions across model families, in the reconstruction task when the compression model is set to LLaMA-2 7B and GPT-3.5. Here R-L stands for Rouge-L and B-S stands for BERTScore. We notice a good generalization of compressions from GPT-3.5 to LLaMA-2 13B.}
\label{tab:general2}
\end{table}

\section{Related Work}
\label{sec:related}
The goal of prompt compression is to make the input of language models shorter, therefore accelerate the inference time and reduce the computational cost.
 \citet{wingate-etal-2022-prompt} and \citet{chevalier-etal-2023-adapting} employ learned soft prompts instead of discrete tokens to compress language model inputs. Soft prompts have been demonstrated to preserve high-level information in context. However, this approach is not suitable for black-box models.

The other line of work compresses prompts by removing redundant discrete tokens. \citet{li-etal-2023-compressing} calculate self-information for each token in a prompt using a smaller LM, then eliminate lexical units with low self-information. Similarly, \citet{jiang-etal-2023-llmlingua} employ a comparable strategy for prompt compression but introduce a novel budget controller. This controller dynamically allocates compression ratios to different prompt components. Our method differs from theirs in two aspects: firstly, our approach is generation-based, harnessing the strong generation capabilities of LMs to produce compressed text. Secondly, while these two methods are task-agnostic, we propose a lightweight framework to adapt LMs to task-specific prompt compressors without training.

Other researchers explore to empower efficient inference for some specific tasks: \citet{xu2023recomp} train extractive and abstractive summarization models to compress retrieved documents. Tailered for the in-contect learning task, \citet{zhou2023efficient} predict the number of examples needed for each sample to avoid redundancy. \citet{yin-etal-2023-read} focus on the role of task definitions in prompts and propose to iteratively remove the unimportant contents in task definitions.

\section{Conclusion}
This work aims to solve the challenge of enhancing prompt compression for diverse tasks. Our preliminary study reveals that styles have an impact in prompt compression. In light of that, enabling the compression model to compress in styles beneficial for different tasks could be a solution for this challenge. We propose Style-Compress, a training-free framework that adapts a smaller LM to compress prompts for a larger LM across different tasks. We first generate and evaluate effective compressions with diverse styles, then use theses examples to guide the compression model. By learning task-relevant styles, our model demonstrates improved performance across four tasks. With limited data for adaptation and queries for the larger LM, Style-Compress is able to compress at compression ratios like 0.25 and 0.5 with minimal or no performance drop on tasks including summarization, multi-hop QA and reasoning.

\section*{Limitations}
A limitation of our method is the difficulty in controlling the length of compressed prompts, a natural drawback of generative methods. Although we implement some procedures like including the target token count in the instructions for compression and setting parameters like min\_new\_token and max\_new\_token when generating with the compression model, we still find it challenging to instruct smaller LMs to generate compressed prompts that adhere to a target compression ratio set by us. In our experiment using a smaller LM, i.e. LLaMA-2 7B as a compression model, we find that the actual compression ratios are usually far lower than the target ratio when the target ratio is a bit higher, e.g. 0.5. Therefore, our method is at a disadvantage compared to extraction-based methods in terms of precise length control.

In addition, since the compression is generated by a language model, we cannot guarantee the generated compression is completely faithful to original prompt. Like other machine-generated text, prompts compressed by this method could suffer from hallucinations \citep{huang2023survey} or bias inherited from the compression LM.

\section*{Acknowledgements}
We thank the reviewers and the area chair. This work was supported by Beijing Science and Technology Program (Z231100007423011) and Key Laboratory of Science, Technology and Standard in Press Industry (Key Laboratory of Intelligent Press Media Technology). Xiaojun Wan and Tianxing He are the corresponding authors.

\bibliography{anthology,custom}

\appendix

\clearpage

\section{Prompt Instructions}
\label{appsec:instrution}
Here we list instructions used for compressing prompts.
\subsection{Instructions for Prompt Compression}
\label{appsec:instrution_compr}

The vanilla instruction for prompt compression is shown in Figure \ref{fig:vanilla-instruct}. When using the instruction one needs to add the corresponding elements inside the angle brackets:

\begin{figure}[h]
    \centering
    \tcbset{width=\linewidth}
    \begin{tcolorbox}
Compress the following text into <target token count> tokens, as such you can still understand the original meaning of it. <style instruction>\\
\textbf{Original Text}: <original prompt>\\
\textbf{Compressed Text}:
\end{tcolorbox}
\caption{Vanilla instruction for prompt compression.}
\label{fig:vanilla-instruct}
\end{figure}

The few-shot instruction for prompt compression is shown in \ref{fig:few-shot-instruct}.

\begin{figure}[h]
    \centering
    \tcbset{width=\linewidth}
    \begin{tcolorbox}
Follow the demonstrations to compress the original text in <target token count> tokens.\\
-------\\
\textbf{Original text:}<original prompt of demonstration 1>\\
\textbf{Compressed text:}<compressed prompt of demonstration 1>\\
-------\\
\textbf{Original text:}<original prompt of demonstration 2>\\
\textbf{Compressed text:}<compressed prompt of demonstration 2>\\
-------\\
......\\
-------\\
\textbf{Original text:}<original prompt>\\
\textbf{Compressed text:}
\end{tcolorbox}
\caption{Few-shot instruction for prompt compression.}
\label{fig:few-shot-instruct}
\end{figure}

\subsection{Instruction for Styles}
\label{appsec:instrution_style}
\begin{itemize}
    \item loc-begin: "Focus on the initial portion of the text."
    \item loc-mid: "Focus on the middle portion of the text."
    \item loc-end: "Focus on the latter portion of the text."
    \item loc-all: "Compress the entire text comprehensively, ensuring all parts are condensed effectively."
    \item style-ab: "Make it more abstractive, by paraphrasing in your own words or restructuring the original text to convey the same meaning in a more concise form."
    \item style-ex: "Make it more extractive, by selecting the most important phrases or sentences to condense the content."
    \item readable: "Make sure the compressed text is fluent, grammatically correct, and human-readable."
    \item unreadable: "Do not make it human-readable. Abuse of language mixing, abbreviations, symbols(unicode and emojis) to aggressively compress it."
    \item format-aware: "If the original text has a specific structure or format, maintain the key sentences from the original to preserve this structure or format."
    \item for reconstruction: "This is for the reconstruction task."
    \item for summarization: "This is for the summarisation task."
    \item for qa: "This is for the multi-hop QA task."
    \item for reasoning: "This is for the reasoning task."
\end{itemize}

\section{Supplementary Ablation Study and Parameter Analysis}
\label{appsec:ablation}

\textbf{Effects of comparative advantage. } When choosing the best compressed prompts as examples, we assess candidates based on their comparative advantages. We have two variants of comparative advantage in this work: $CA^\text{min}$ and $CA^\text{mid}$. To examine the effects of comparative advantage, we compare these variants against using absolute advantage. In the absolute advantage setting, different compressed prompts of the same original ones are not compared against each other. Instead, the compression LM processes a new original prompt each time, and all compression samples are compared together using the absolute value from the metrics. The iteration count and computational complexity are kept the same for a fair comparison.

The results of this ablation study, shown in Table \ref{tab:ablation_ca}, demonstrate that comparative advantage outperforms the absolute version. $CA^\text{mid}$ performs better for reasoning tasks, while $CA^\text{min}$ is more effective for reconstruction tasks. The reasoning task uses a discrete metric for each sample (either 1 or 0). In this case, the max-min variant often results in various compressed prompts tying for the top spot with the same comparative advantage (which is 1), whereas the max-mid variant can better distinguish the best compression samples.

\begin{table}[]
\resizebox{\columnwidth}{!}{%
\begin{tabular}{@{}l|l|ccll@{}}
\toprule
            & Reasoning     & \multicolumn{4}{c}{Reconstruction}                                                           \\ \cmidrule(l){2-6} 
            & \multicolumn{1}{c|}{Acc.}& Rouge1         & Rouge2         & \multicolumn{1}{c}{RougeL} & \multicolumn{1}{c}{BS} \\ \midrule
absolute   & 0.107         & 0.311          & 0.212          & 0.242                      & 0.867                         \\
$CA^\text{min}$ & 0.115         & \textbf{0.372} & \textbf{0.274} & \textbf{0.297}             & \textbf{0.88}                 \\
$CA^\text{mid}$ & \textbf{0.12} & 0.346          & 0.242          & 0.267                      & 0.872                         \\ \bottomrule
\end{tabular}%
}
\caption{Ablation study of comparative advantage.}
\label{tab:ablation_ca}
\end{table}

\textbf{Effect of iteration count. } Here we analyse how  $M$ and $N$ (including $N_{style}$ and $N_{icl}$) affect the compression performance. As shown in Table \ref{tab:ablation_mn}, accuracy of our method on the reasoning task becomes higher as $M$ increases: compared to 10, accuracy increased by 0.5\% when $M$ is set to 20, and 1.2\% when $M$ is set to 50. On the reconstruction task, we do not observe a clear improvement when $M$ is increased from 10 to 50. In addition, the optimal setting of iteration count $N_{style}$ and $N_{icl}$ are 5 for both tasks.

\begin{table}[]
\resizebox{\columnwidth}{!}{%
\begin{tabular}{@{}l|l|llll@{}}
\toprule
\multicolumn{1}{c|}{} & Reasoning      & \multicolumn{4}{c}{Reconstruction}                                \\ \midrule
M   & \multicolumn{1}{c|}{Acc.} & \multicolumn{1}{c}{Rouge1} & \multicolumn{1}{c}{Rouge2} & \multicolumn{1}{c}{RougeL} & \multicolumn{1}{c}{BS} \\ \midrule
5                     & 0.107          & 0.233          & 0.105          & 0.152          & 0.847          \\
10                    & 0.12           & 0.372          & \textbf{0.274} & \textbf{0.297} & 0.88           \\
20                    & 0.125 & 0.352          & 0.259          & 0.284          & \textbf{0.882} \\
50                    & \textbf{0.132}           & \textbf{0.379} & 0.252          & 0.277          & 0.873          \\ \midrule
N/2 & \multicolumn{1}{c|}{Acc.} & \multicolumn{1}{c}{Rouge1} & \multicolumn{1}{c}{Rouge2} & \multicolumn{1}{c}{RougeL} & \multicolumn{1}{c}{BS} \\ \midrule
3                     & 0.109          & 0.343          & 0.188          & 0.235          & 0.864          \\
5   & \textbf{0.12}             & \textbf{0.372}             & \textbf{0.274}             & \textbf{0.297}             & \textbf{0.88}          \\
10                    & 0.111          & 0.356          & 0.257          & 0.281          & 0.876          \\ \bottomrule
\end{tabular}%
}
\caption{Effect of iteration count, $M$ and $N$. Here $N/2 = N_{style} = N_{ICL}$.}
\label{tab:ablation_mn}
\end{table}



\begin{figure}
    \centering
    \includegraphics[width=0.9\columnwidth]{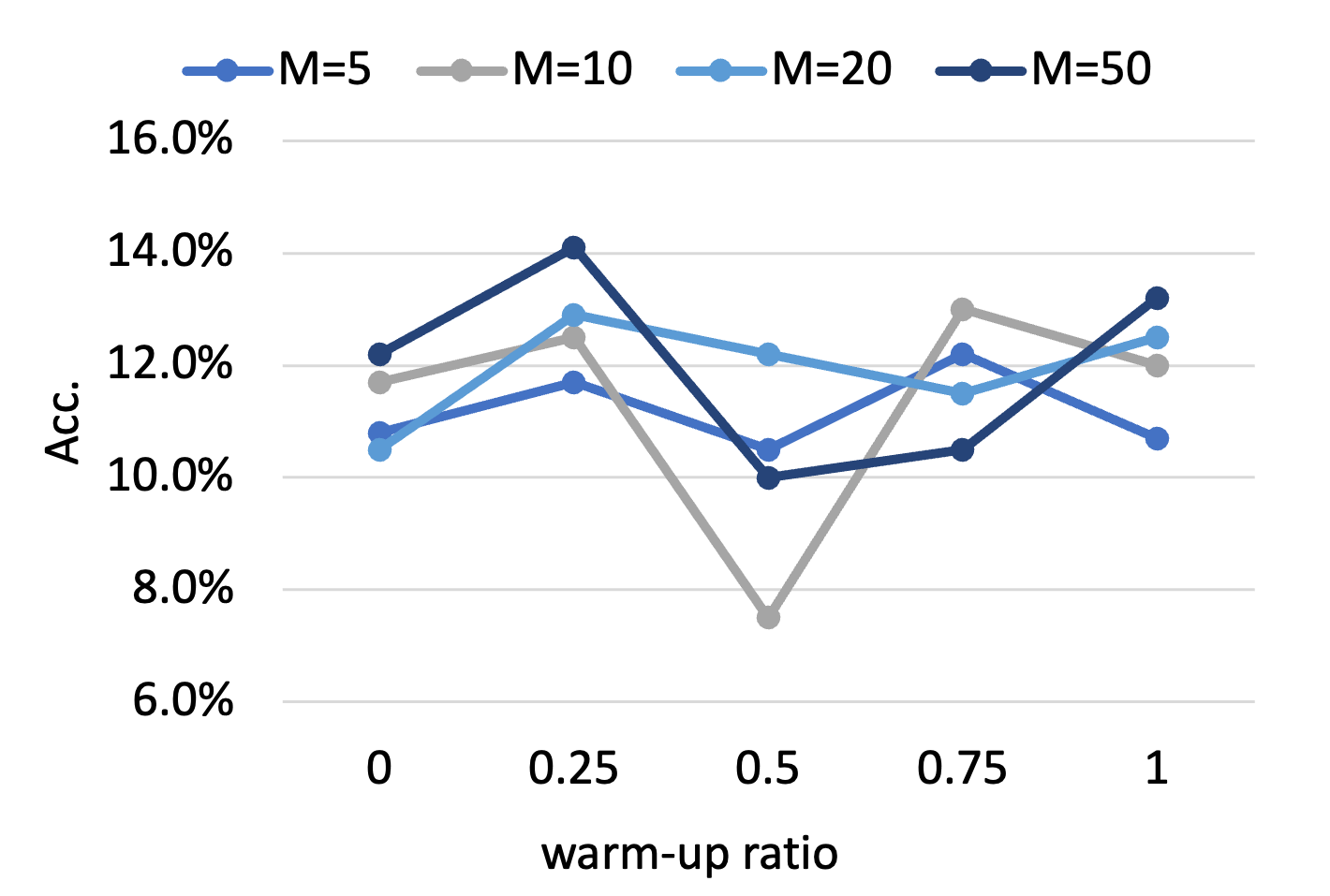}
    \caption{Effect of warm-up ratio in style controller. When ratio is set to 1, style controller randomly samples styles all the time.}
    \label{fig:ablation_warmup}
\end{figure}

\textbf{Effect of warming-up in style controller.} In Figure \ref{fig:ablation_warmup}, we test our method with different warm-up ratios when $M$ is set to 5, 10, 20, and 50. The results demonstrate that if we use weighted random sampling from the very beginning, the performance across different values of $M$ is similar. As the warm-up ratio increases to 0.25, accuracy improves for all values of $M$, with more iterations leading to greater improvements. However, as the ratio continues to increase, we observe a performance drop at a ratio of 0.5, followed by continuous improvement up to a ratio of 1.

\section{Details of Tasks and Datasets}
\label{appsec:tasks}
In the original prompt reconstruction task, the evaluation LLM is asked to reconstruct the compressed prompt to its original version. Higher similarity between the reconstruction of the compressed prompt and the original prompt indicates more effective compression. We use the BBC News dataset \citep{greene06icml} for this task.

For the text summarization task, we use the CNN/DailyMail dataset \citep{see-etal-2017-get}. We compare the summaries generated from the compressed text with the human-written reference summaries. For the reconstruction and summarization tasks, we compress news articles in the dataset.

Multi-hop QA is a challenging question-answering task that requires language models to extract and combine multiple pieces of information from the context and perform multi-hop reasoning to answer the question. We use the HotpotQA dataset \citep{yang-etal-2018-hotpotqa} for this task, where LMs are tasked with answering questions based on 10 pieces of Wikipedia articles. In this task, we concatenate the provided Wikipedia articles for each question and then compress them.

In the CoT reasoning task, LMs are instructed with examples that include the question, intermediate reasoning steps, and the final answer, to answer another math question. CoT prompting has been proven to enhance LMs' complex reasoning capabilities \citep{wei2022chain}. We use GSM8k \citep{cobbe2021gsm8k}, a dataset of human-written grade school math word problems for this task. We only compress the intermediate reasoning steps of each example, and then concatenate it with the corresponding question and final answer together as a demonstration provided to LLMs.

For each dataset, we take 200 samples as the test set and additionally sample 10-100 samples for the adaptation phase (according to different settings). For the samples to be compressed, we preprocess them to be within 1000 tokens, with any excess parts being truncated.

\section{Examples of Compressed Prompts}
\label{appsec:examples}
Here we present some examples of prompt compression for the reconstruction task (as shown in Figure \ref{fig:example1}) and CoT reasoning task (the original prompt shown in Figure \ref{fig:example2} and the compressed shown in Figure \ref{fig:example3}). When compressing for the reconstruction task on the BBC dataset, it can be observed that Style-Compress learns a summary-like fluent style. Therefore, when we use GPT-3.5 to reconstruct the compressed content, the difference between the original and recovered texts is minimal due to this fluency. Compared with this, the vanilla method adopts a fact-listing format, and the text compressed by LLM-Lingua is hard for humans to understand. After recovery with GPT-3.5, both methods perform worse in metrics compared to ours. When Style-Compress is applied to the reasoning task, we observe a precise and logical style in the compressed prompt. We compare its effectiveness in condensing demonstrations with few-shot CoT prompting, demonstrating that our compression method enhances accuracy on the math reasoning task compared to zero-shot prompting.

\begin{figure*}
    \centering
    \tcbset{width=\linewidth}
    \begin{tcolorbox}
\textbf{Original Prompt:}
Boris Johnson has submitted evidence to MPs investigating whether he misled Parliament over Covid rule-breaking parties in Downing Street. The former prime minister, who denies misleading MPs, wants his defence to be published as soon as possible. The Privileges Committee said it was reviewing the material to make redactions before publication. Mr Johnson(......)An investigation by senior civil servant Sue Gray found widespread rule-breaking had taken place and a police inquiry led to 83 people, including Mr Johnson himself, being fined for attending events. The Privileges Committee's initial report said it had seen evidence that "strongly suggests" Covid rule breaches would have been "obvious" to Mr Johnson. \\

\textbf{Compressed by Style-Compress:} Boris Johnson has submitted evidence to MPs investigating whether he misled Parliament over Covid rule-breaking parties in Downing Street. The former prime minister, who denies misleading MPs, wants his defence to be published as soon as possible. The Privileges Committee is reviewing the material to make redactions before publication. Mr Johnson will be questioned by the cross-party committee in a televised session
(77 tokens)

\textbf{Reconstructed by ChatGPT:} Boris Johnson has submitted evidence to MPs investigating whether he misled Parliament over Covid rule-breaking parties in Downing Street. The former prime minister, who denies misleading MPs, wants his defence to be published as soon as possible. The Privileges Committee is reviewing the material to make redactions before publication. Mr Johnson will be questioned by the cross-party committee in a televised session.

\textit{(Rouge-1: 0.199    Rouge-2: 0.190    Rouge-L: 0.196    BERTScore: 0.885)}\\

\textbf{Compressed by Vanilla:} 1. Nicola Sturgeon is SNP leader and her husband Peter Murrell is CEO. 2. Peter Murrell loaned SNP £107,620 to help with cash flow. 3. SNP repaid £47,620 of the loan by Oct 2021. 4. Loan was not initially thought to be a reporting obligation. 5. Electoral Commission records show
(72 tokens)

\textbf{Reconstructed by ChatGPT:}Boris Johnson has submitted evidence to MPs who are investigating whether he misled Parliament. The evidence will be published soon, with redactions made to protect the identities of witnesses. Johnson is scheduled to be questioned in a televised session on Wednesday. If he is found guilty, he could face punishments such as issuing an apology or suspension from Parliament. Some of Johnson's allies have expressed concerns about the independence of the inquiry. Members of Parliament will have a free vote on any sanctions that may be imposed. The committee conducting the investigation is chaired by Labour's Harriet.

\textit{(Rouge-1: 0.205    Rouge-2: 0.105    Rouge-L: 0.142    BERTScore: 0.866)}\\

\textbf{Compressed by LLM-Lingua:} B has he over Covidbreaking in denies, his beTheges makeactions the a an Mr Parliament
The and wouldas is possible the of making appropriate some witnesses
 could be in his
 will was
 any but punishments from his
 say theThe theor, a by committee by breakingid be Westminster Johnson has integrity and ofid servant toobvious (78 tokens)
 
 \textbf{Reconstructed by ChatGPT:} Mr Johnson denies breaking any rules over Covid, as his actions in Parliament make it possible for appropriate witnesses to be in his will. However, the committee by Westminster could say that he is breaking the integrity and obvious servant of the law.
 
 \textit{(Rouge-1:  0.131    Rouge-2: 0.028    Rouge-L: 0.080    BERTScore: 0.819)}
\end{tcolorbox}
    \caption{Examples of different methods compressing the same original prompt for the reconstruction task.}
    \label{fig:example1}
\end{figure*}

\begin{figure*}
    \centering
    \tcbset{width=\linewidth}
\begin{tcolorbox}
\textbf{CoT 3-shot Prompting}\\
\textit{Refer to the following examples to answer the math problem.}

\textit{Example 1}\\
Question:Stanley was standing on the side of the mountain, at an elevation of 10,000 feet, when a gust air blew the comb out of his pocket, causing the comb to fall 4,000 feet to a ledge below. Stanley's brother, Oliver, was also on the mountain, but he was at an elevation of 3,000 feet. Stanley called Oliver on his cellphone and asked Oliver to find Stanley's comb. How many feet must Oliver climb to reach the elevation of Stanley's comb?\\
Answer:The elevation of Stanley's comb was 10,000 - 4,000 = <<10000-4000=6000>>6,000 feet. Thus Oliver, who was at an elevation of 3,000 feet, must climb 6,000 - 3,000 = <<6000-3000=3000>>3,000 feet to reach the elevation of Stanley's comb. The answer is: 3000\\
\textit{Example 2}\\
Question: Carl has four times as many marbles as Sean and Sean has half as many marbles as Cal. If Sean has 56 marbles, how many marbles do Carl and Cal have combined?\\
Answer:Cal has 56*2 = <<56*2=112>>112 marbles. Carl has 4*56 = <<4*56=224>>224 marbles. Carl and Cal have 112+224 = <<112+224=336>>336 marbles combined. The answer is: 336

\textit{Example 3}\\
Question:Paul is at a train station and is waiting for his train. He isn't sure how long he needs to wait, but he knows that the fourth train scheduled to arrive at the station is the one he needs to get on. The first train is scheduled to arrive in 10 minutes, and this train will stay in the station for 20 minutes. The second train is to arrive half an hour after the first train leaves the station, and this second train will stay in the station for a quarter of the amount of time that the first train stayed in the station. The third train is to arrive an hour after the second train leaves the station, and this third train is to leave the station immediately after it arrives. The fourth train will arrive 20 minutes after the third train leaves, and this is the train Paul will board. In total, how long, in minutes, will Paul wait for his train?\\
Answer:The first train stays in the station for 10 minutes * 2 = <<10*2=20>>20 minutes. So Paul waits for the first train for 20 + 10 = <<20+10=30>>30 minutes. The second train arrives after half an hour, which is the same as 60 minutes  2 = <<60/2=30>>30 minutes. It then stays in the station for a quarter of the time the first train did which is 20 minutes 4 = <<20/4=5>>5 minutes. So Paul waits for the second train for 30 + 5 = <<30+5=35>>35 minutes. The third train arrives an hour later, which is the same as 60 minutes * 1 hour = <<60*1=60>>60 minutes. After the final wait, Paul has waited a total of 30 minutes for the first train + 35 minutes for the second train + 60 minutes for the third train + 20 minutes final wait = <<30+35+60+20=145>>145 minutes. The answer is: 145\\
\textit{Question:} Ted the T-Rex was planning to bring potato salad to the dinosaur picnic. He knows that an adult dinosaur will eat 10 lbs of potato salad, and a child will eat half as much as an adult. If there will be 20 adults and 5 children at the picnic, how many pounds of potato salad does Ted need to bring to the picnic if he hopes to have enough to feed everyone?\\
\\
Answer: An adult will eat 10 lbs of potato salad and a child will eat half as much which is 5 lbs. Therefore, the total amount of potato salad needed to feed all the adults and children is: 10 lbs x 20 adults = <<10*20=200>>200 lbs + 5 lbs x 5 children = <<5*5=25>>25 lbs. Ted needs to bring a total of 200 + 25 = <<200+25=225>>225 lbs of potato salad to the picnic. The answer is 225.
(Correct!)
\\

\end{tcolorbox}
\caption{}
    \label{fig:example2}
\end{figure*}

\begin{figure*}
    \centering
    \tcbset{width=\linewidth}
    \begin{tcolorbox}
\textbf{Style-Compressed 3-shot CoT Reasoning}

\textit{Refer to the following examples to answer the math problem.}

\textit{Example 1}\\
Question: Stanley was standing on the side of the mountain, at an elevation of 10,000 feet, when a gust air blew the comb out of his pocket, causing the comb to fall 4,000 feet to a ledge below. Stanley's brother, Oliver, was also on the mountain, but he was at an elevation of 3,000 feet. Stanley called Oliver on his cellphone and asked Oliver to find Stanley's comb. How many feet must Oliver climb to reach the elevation of Stanley's comb?\\
Answer: Stanley's comb elevation: 6,000 ft (<<6000=3000>>3,000 ft climb to reach). 
The answer is: 3000\\

\textit{Example 2}\\
Question:Carl has four times as many marbles as Sean and Sean has half as many marbles as Cal. If Sean has 56 marbles, how many marbles do Carl and Cal have combined?\\
Answer:Cal has 112 marbles. Carl has 224 marbles.
The answer is: 336\\

\textit{Example 3}\\
Question:Paul is at a train station and is waiting for his train. He isn't sure how long he needs to wait, but he knows that the fourth train scheduled to arrive at the station is the one he needs to get on. The first train is scheduled to arrive in 10 minutes, and this train will stay in the station for 20 minutes. The second train is to arrive half an hour after the first train leaves the station, and this second train will stay in the station for a quarter of the amount of time that the first train stayed in the station. The third train is to arrive an hour after the second train leaves the station, and this third train is to leave the station immediately after it arrives. The fourth train will arrive 20 minutes after the third train leaves, and this is the train Paul will board. In total, how long, in minutes, will Paul wait for his train?\\
Answer:1. First train stays 20 min 2. Paul waits for first train 30 min 3. Second train arrives after half an hour 30 min 4. Second train stays 5 min 5. Third train arrives 1 hour later 60 min 6. Paul waits total 145 min. 
The answer is: 145\\
\textit{Question: }Ted the T-Rex was planning to bring potato salad to the dinosaur picnic. He knows that an adult dinosaur will eat 10 lbs of potato salad, and a child will eat half as much as an adult. If there will be 20 adults and 5 children at the picnic, how many pounds of potato salad does Ted need to bring to the picnic if he hopes to have enough to feed everyone?\\
\\
Answer:
20 adults x 10 lbs/adult = 200 lbs 5 children x 5 lbs/child = 25 lbs Total pounds of potato salad = 225 lbs The answer is 225.\\
(Correct!)\\
\\
\textbf{Zero-shot Prompting}\\
\textit{Answer the following math problem.}\\
\textit{Question: }Ted the T-Rex was planning to bring potato salad to the dinosaur picnic. He knows that an adult dinosaur will eat 10 lbs of potato salad, and a child will eat half as much as an adult. If there will be 20 adults and 5 children at the picnic, how many pounds of potato salad does Ted need to bring to the picnic if he hopes to have enough to feed everyone?\\
Answer:  35 pounds
(Wrong!)\\

\end{tcolorbox}
\caption{An example of our method, Style-Compress, for the reasoning task. We make a comparison between zero-shot or few-shot CoT prompting on the reasoning task, and shows that our compression method can condense the demonstrations will increase accuracy than zero-shot prompting. The responses are generated by LLaMA-2 13B.}
\label{fig:example3}
\end{figure*}
\end{document}